\documentclass[runningheads,a4paper]{llncs}

\usepackage{enumitem}
\setlist{nolistsep}

\usepackage{amssymb}
\usepackage{graphicx}
\usepackage[usenames,dvipsnames]{color}

\usepackage{url}

\usepackage{amsmath}
\usepackage{textcomp} 
\usepackage{xspace} 
\usepackage{bbding} 
\usepackage{pifont} 
\usepackage{stmaryrd} 
\usepackage{mdwlist} 
\usepackage[cp1252]{inputenc}

\begin{document}
\newcommand{\epar}[1]{``#1''}
 
\authorrunning{PREPRINT}

\title{Towards Software Development\\ For Social Robotics Systems
} 
\author{Chong Sun, Jiongyan Zhang, Cong Liu, Barry Chew Bao King,\\ Yuwei Zhang, Matthew Galle, Maria Spichkova}
\institute{RMIT University,  Australia \\  \email{$\{$s3557753, s3589957, s3556054, s3584485, s3492095, s3491364$\}$@student.rmit.edu}, 
\email{maria.spichkova@rmit.edu.au} }
\maketitle

\begin{abstract}
In this paper we introduce the core results of the project on 
software development for social robotics systems. 
The usability of maintenance and control features is crucial for many kinds of systems, but in the case of social robotics we also have to take into account that (1) the humanoid robot physically interacts with humans, (2) the conversation with children might have different requirements in comparison to the conversation with adults. 
The results of our work were implement for the humanoid PAL REEM robot, but their core ideas can be applied for other types of humanoid robots.
We developed a web-based solution that supports the management of robot-guided tours, provides recommendations for the users as well as allows for a visual analysis of the data on previous tours.
\end{abstract}

\section{Introduction}

Social robotics is an emerging research area. Over the last years there were many publications on application of robotics for healthcare and rehabilitation,  household and service, healthcare and rehabilitation, companionship, etc., cf. \cite{yumakulov2012imagery}. 
The core function of social robots is  assisting people through social interaction, in many cases involving also a physical interaction.
A highly cited\footnote{462 citations according to the Google Scholar, retrieved 20 December 2017} paper of Feil-Seifer and Mataric \cite{feil2005defining} defines the concept of socially assistive robotics.
Another highly cited\footnote{592 citations according to the Google Scholar, retrieved 20 December 2017} paper of Duffy \cite{duffy2003anthropomorphism} discusses the use of anthropomorphic paradigms to augment the functionality and behavioural characteristics of a robot use of human-like features for social interaction with people.  
 
To understand the impact and capabilities of robots on the future of work, it is crucial to identify, observe and measure interactions between robots and humans, as well as to develop systems that support these observations and measurement.
The focus of our project is on social robotics: analysis of interaction between humans and humanoid robots, as well as the corresponding support in the development and maintenance of humanoid robotics systems that are acting autonomously.  
 
The work was conducted in collaboration 
with Commonwealth bank (CBA) under support of the Australian Technology Network (ATN). 
This project was a part of the ATN CBA Robotics Education and Research program, and continued our previous research on the topic of social robotics using a humanoid PAL REEM robot to conduct the experiments: 
The first project was dedicated to the development of a general framework for a REEM guided tour as well as its implementation for the REEM robot, cf. \cite{clunne2017modelling}. 

\emph{Contributions:} The current project extends 
the developed framework by new features, such as (1) providing a web-based application for navigating the robot during the phase of collecting the spatial information, (2) creating and editing the tour files in a user-friendly manner, 
(3) providing recommendations for the users,  as well as (4) allowing for a visual analysis of the data on previous tours.
In this paper we present a solution that allows lab assistants to interact with the robotics system without having any technical knowledge about the system and can be operated by any exhibition, or lab staff member or social psychologist.

The project we present in this paper is a part of the RMIT University activities on enhancing learning experience by collaborative industrial projects \cite{christianto2017software,iceer_projects,spichkova2017autonomous,spichkova2015formal,spichkova2016formal}. 
The core results of our previous project on social robotics are presented in \cite{clunne2017modelling}: We focused on the Lab tours use case, where
the robot takes guests on tours of our Innovation Labs and answers related questions. The framework presented in \cite{clunne2017modelling} is based on a formal framework for modelling and analysis of autonomous systems and their compositions \cite{SpichkovaSimic2015}, and  can be applied to any kind of guided tours, as changing the application domain would mean changing only on the content of information provided about exhibits. In the project we present in this paper, we went further to extend the framework with web-based interface providing many useful features. While the old version with the voice commands is more human-oriented, the new web interface can be useful for a noisy environment. 
Thus, in some cases the spatial information for the tours has to be collected in noisy environments where the identification of voice commands can be difficult or even compromised. This might happen, for example, when the exhibition construction works are in progress, and waiting till all works are finished might led to additional delays. 
 
\emph{Outline:} The rest of the paper is organised as follows. 
Section~\ref{sec:related} presents related work. 
The architecture of the developed system as well as its core functionalities  are introduced in Section~\ref{sec:arch}.  
Section~\ref{sec:conclusions} summarises the paper and introduces directions of our future work.

\section{Related Work}
\label{sec:related}

Duffy et al. \cite{duffy1999social}  presented the concept of Social Robot Architecture, which integrates the key elements of agenthood and robotics in a coherent and systematic manner. 
 
 The ethical and social implications of robotics were discussed by Lin et al. in \cite{lin2011robot}.
 Young  et al. \cite{young2009toward}  examined social-psychology concepts to apply them to the human-robot interaction. 
 
Eyssel et al. \cite{eyssel2012if} presented a case study where they analysed the  effects of robot features (human-likeness and gender) and user characteristics on the human-robot interaction acceptance and psychological anthropomorphism. 
Salem et al. \cite{salem2011effects} analysed the effects of gesture on the perception of psychological anthropomorphism, by conducting a case study using the Honda humanoid robot. 
Trovato et al. \cite{trovato2012cross} conducted a 
 cross-cultural study on generation of culture dependent facial expressions of humanoid  robot. 
Sabanovic et al. \cite{sabanovic2006robots} discussed the use of observational studies of human-robot social interaction in open human-inhabited environments.  
Klein  and Cook \cite{klein2012emotional} analysed and compared the findings  in the UK and Germany on robot-therapy with emotional robots as a treatment approach for people with cognitive impairments. 

 There were also a number of surveys and literature reviews on the related topics.
 A survey on social robots for long-term interaction was presented in~\cite{leite2013social}.
A systematic review on application of social robotics in the Autism Spectrum Disorders treatment was presented in~\cite{pennisi2016autism}. 
Cabibihan et al.~\cite{cabibihan2013robots} presented a survey on the roles and benefits of social robots in the therapy of children with autism.

Alemi et al. \cite{alemi2015impact} examined the effect of robot assisted language learning  on the anxiety level and attitude in English vocabulary acquisition amongst Iranian EFL junior high school students. The results demonstrated that application of social robotics in this context can increase learners' engagement as well as satisfaction from the education process.
Shimada et al. \cite{shimada2012can} used a social robot as a teaching assistant in a class for children's collaborative learning, and concluded that a robot can increase children's motivation of the class, but cannot increase their learning achievement.

Glas et al. \cite{glas2012interaction} introduced a design framework
enabling the development of social robotics applications by cross-disciplinary teams
of programmers and interaction designers.

\section{System Architecture and core features}
\label{sec:arch}

\begin{figure}[ht!]
\begin{center}
\includegraphics[width=10cm]{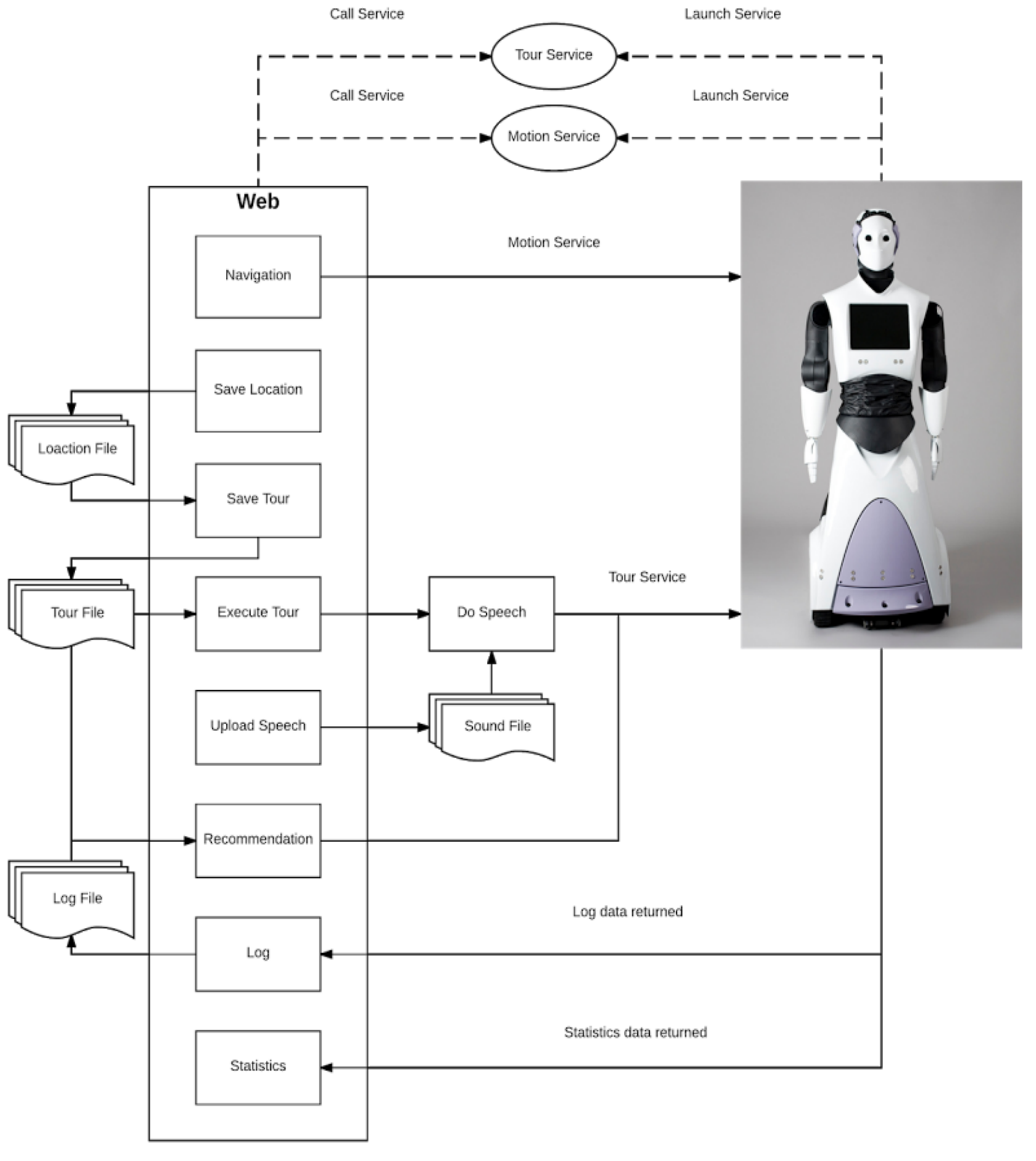}
\end{center}
\caption{System Architecture}
\label{fig:arch}
\end{figure}

\begin{figure}[ht!]
\begin{center}
\includegraphics[width=12cm]{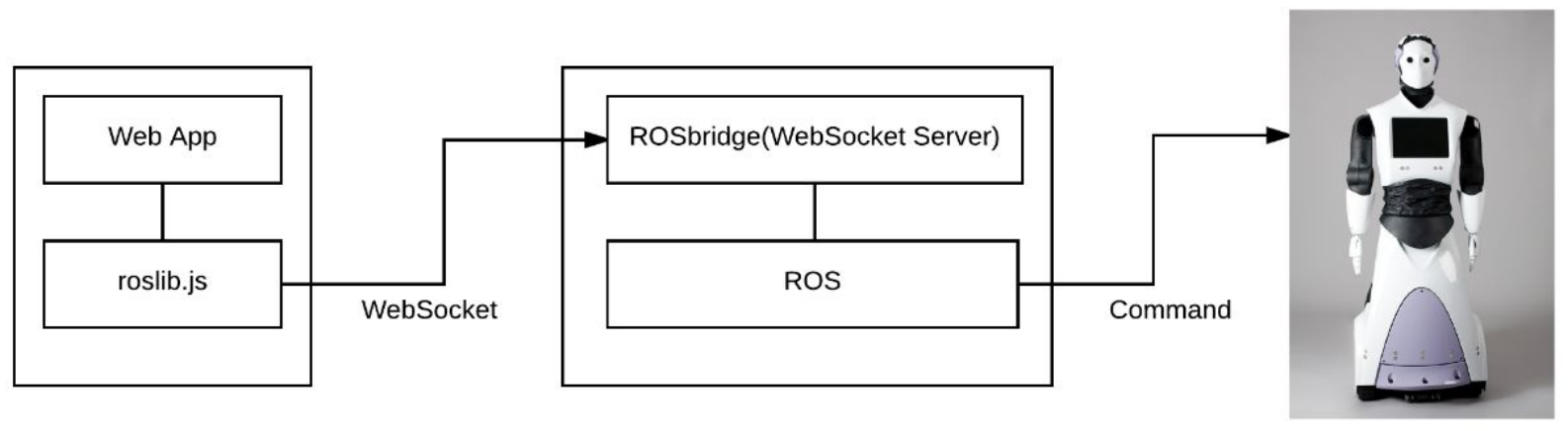}
\end{center}
\caption{System Communication}
\label{fig:com}
\end{figure}

The architecture of the proposed system is demonstrated in Figure~\ref{fig:arch}.  The core 
 physical component of the system is the REEM robot on which the Robotics Operating System (ROS) is running to enable precise control from high-level programs.
ROS provides services for Web-Ros communication, cf. Figure~\ref{fig:com}: 
ROS  side can launch a service
  while the web-interface  can call a service. 
  In this project, we focused on the Tour   and Motion Services. 

Like in our previous project \cite{clunne2017modelling}, our work was divided into two phases: 
\begin{itemize}
	\item
	Phase 1 was conducted in the RMIT University VXLab   (Melbourne, Australia). The introduction to the VXLab facilities can be found in
\cite{blech2014cyber,blech2015visualization,spichkova2013abstract}. 
	The web-based interface was developed using  a simulated environment provided by a ROS robot software development framework. 	
	\item
	Phase 2 was conducted in the CBA Labs (Sydney, Australia).
	A number of experiments were conducted to apply the developed web-based interface to a real REEM robot and to simulate the scenario when an operator prepares an exhibition/lab tour and executes it, both in simulated environment and on a real robot.
\end{itemize}

To develop the web-based interface, we applied React.js, an open-source JavaScript library. To execute the JavaScript code server-side we applied Node.js, an open-source JavaScript runtime environment. 
Node.js provided the management of dependencies to certain web packages that were
 required for certain features to be used such as UI elements and the ROS-bridge API.
 
  ROS (Hydro Medusa) was used as the robot operating system that provides
 interfaces to the REEM robot's sensors, motors, actuators and speakers, by utilising Python and C++ libraries. The robot gesture, movement and navigation functionalities relied on ROS libraries. 
 The experiments were conducted under the Ubuntu 12.04  platform.
 The Gazebo simulator 2.2.3 was used as the simulation environment to test the
 capabilities of the robot.   During simulation,
  the movements of the robot was portrayed through control of the RViz visualisation.
 
  Converting text to speech was conducted using 
  \begin{itemize}
  \item IBM  Watson text-to-speech (TTS) service for the simulation, and 
  \item  on-board Acapela TTS for the experiments on the real robot, 
  \end{itemize}  
  In the simulation environment, TTS relied on generating wave files
  with Watson, then playing them back through any sound player. When deploying on the  actual robot this process is handled by Acapela, a TTS engine from Acapela Group.

\begin{figure}[ht!]
\begin{center}
\includegraphics[width=7.5cm]{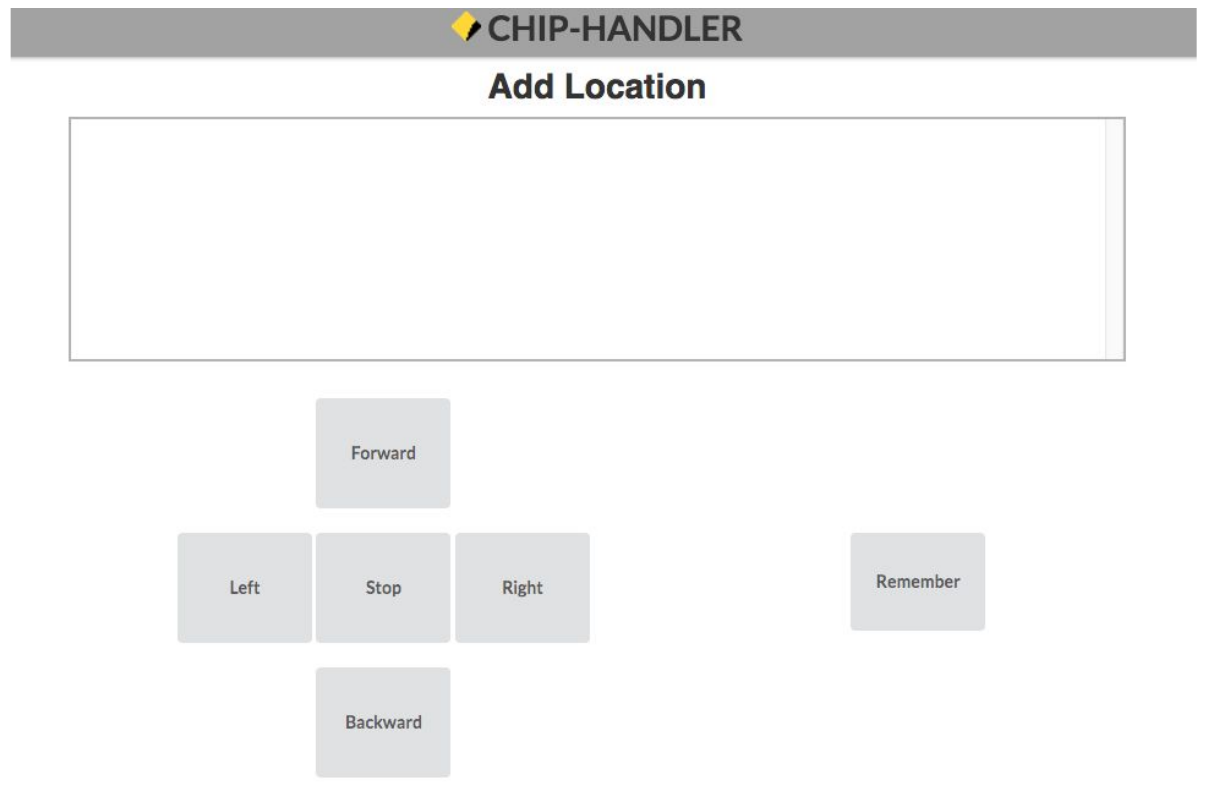}
\end{center}
\caption{Navigation management}
\label{fig:control}
\end{figure}

Figure \ref{fig:control} presents the control page, where the movements of the robot can be controlled by using the corresponding menu items. This provides the functionality necessary to create the lab tours: to
navigate the robot, and to store the current locations of the robot.\\
~

\begin{figure}[ht!]
\begin{center}
\includegraphics[width=7.5cm]{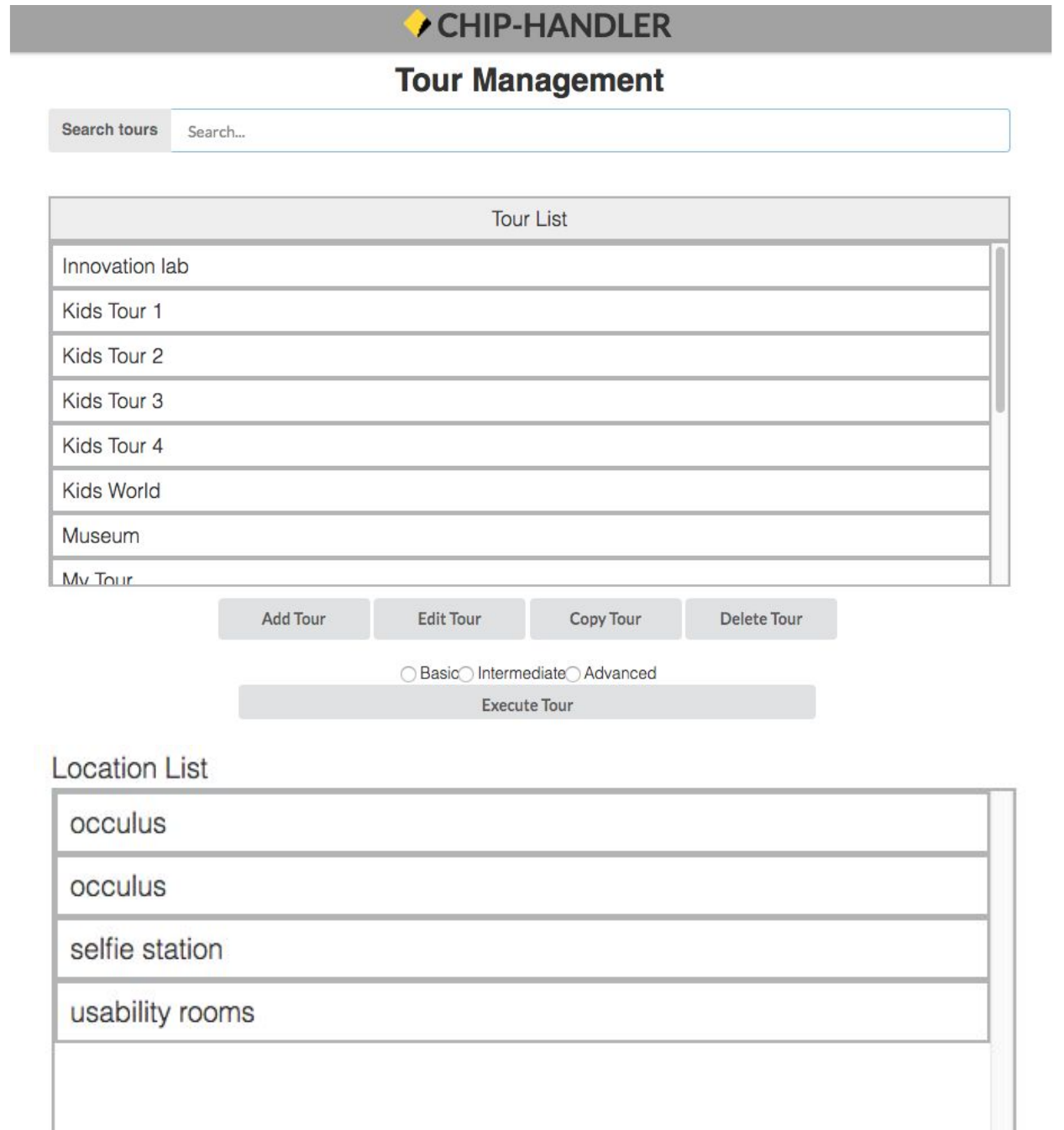}
\end{center}
\caption{Tour Management}
\label{fig:manage}
\end{figure}

\begin{figure}[ht!]
\begin{center}
\includegraphics[width=7.5cm]{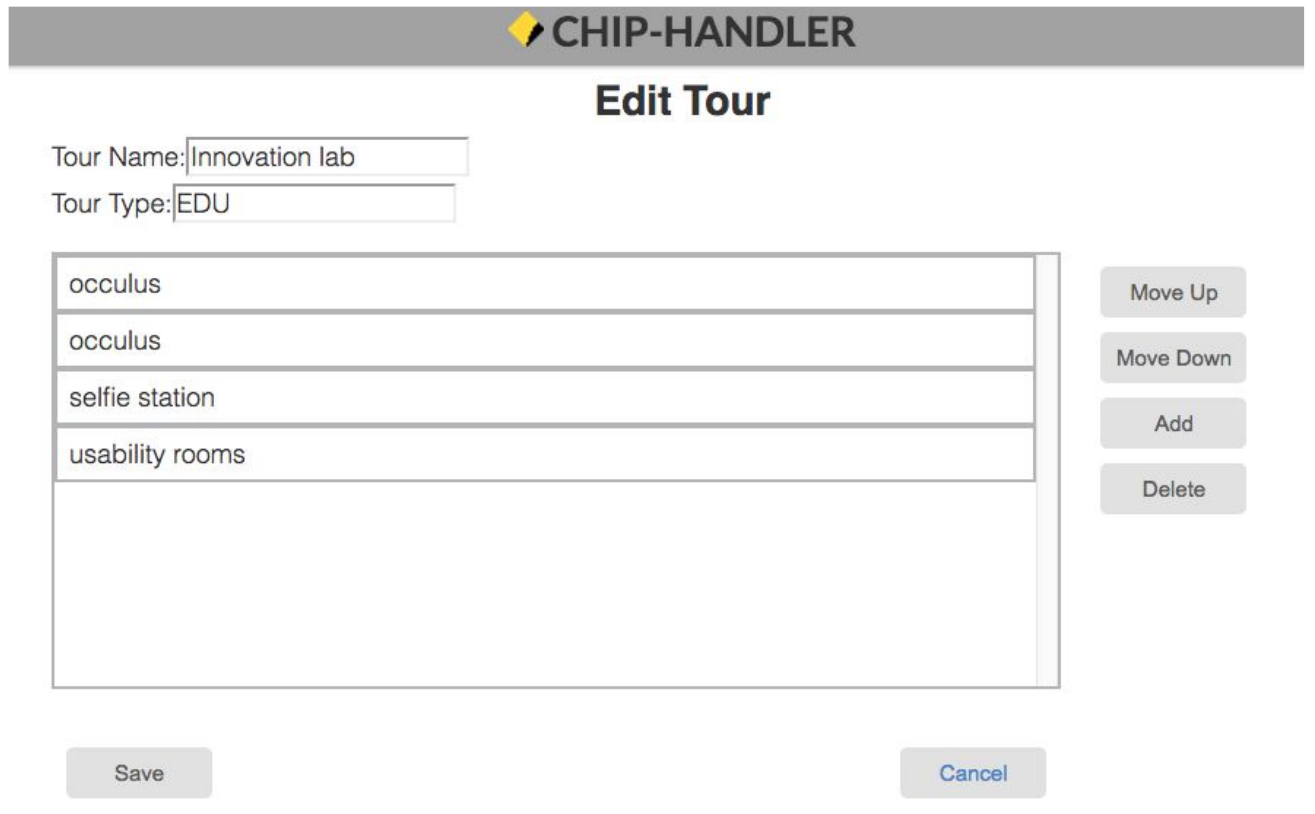}
\end{center}
\caption{Functionality to edit the content of a tour}
\label{fig:edit}
\end{figure}

\newpage
Figure \ref{fig:manage} presents the management page to display all the existing tours and all the locations within each tour, as well as to manage them. 
We can edit the content of an existing  tour (cf. Figure~\ref{fig:edit}), add new tours, copy existing tours, search for particular tours or locations, etc. 

\begin{figure}[ht!]
\begin{center}
\includegraphics[width=7.5cm]{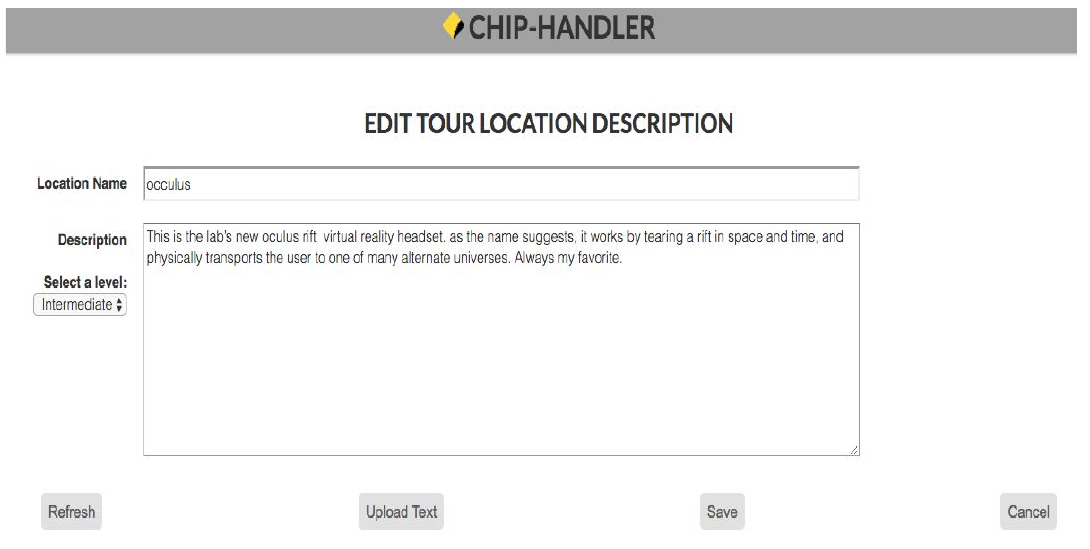}
\end{center}
\caption{Functionality to edit the information on the \emph{occulus} location}
\label{fig:edit_loc}
\end{figure}

Figure \ref{fig:edit_loc} demonstrates the functionality to edit the information on a location named \emph{occulus}. 
The text within the description field will be a part of the speech within the guided tour: the text-to-speech module of the robot system will transform the text to the speech when the robot approaches the location.

\begin{figure}[ht!]
\begin{center}
\includegraphics[width=7.5cm]{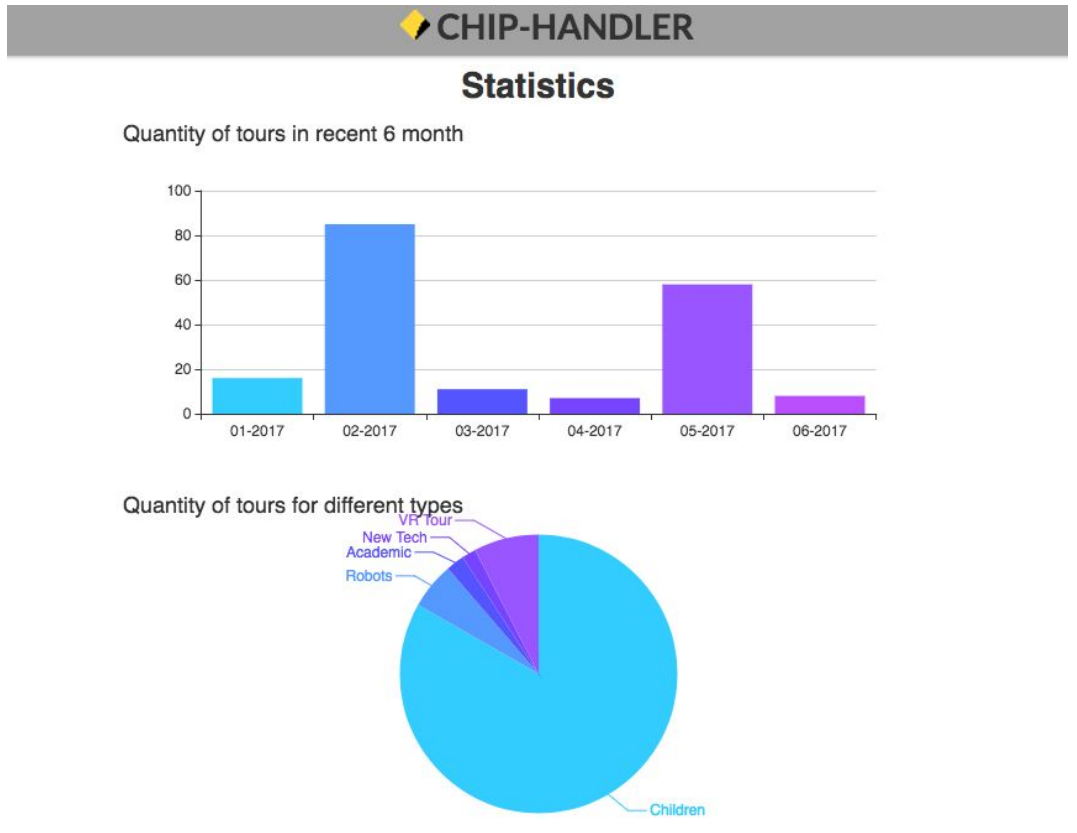}
\includegraphics[width=7.5cm]{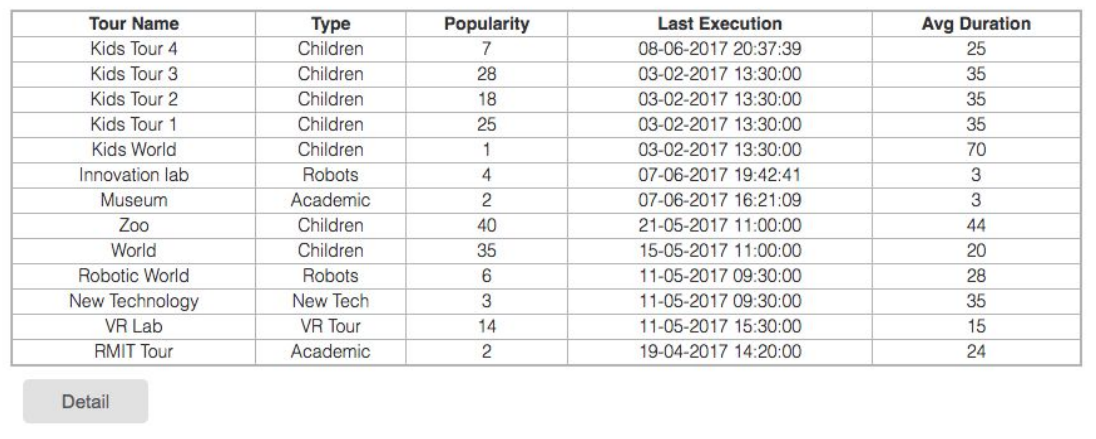}
\end{center}
\caption{Analysis of the data on the previous tours}
\label{fig:stat1}
\end{figure}

\begin{figure}[ht!]
\begin{center} 
\includegraphics[width=7.5cm]{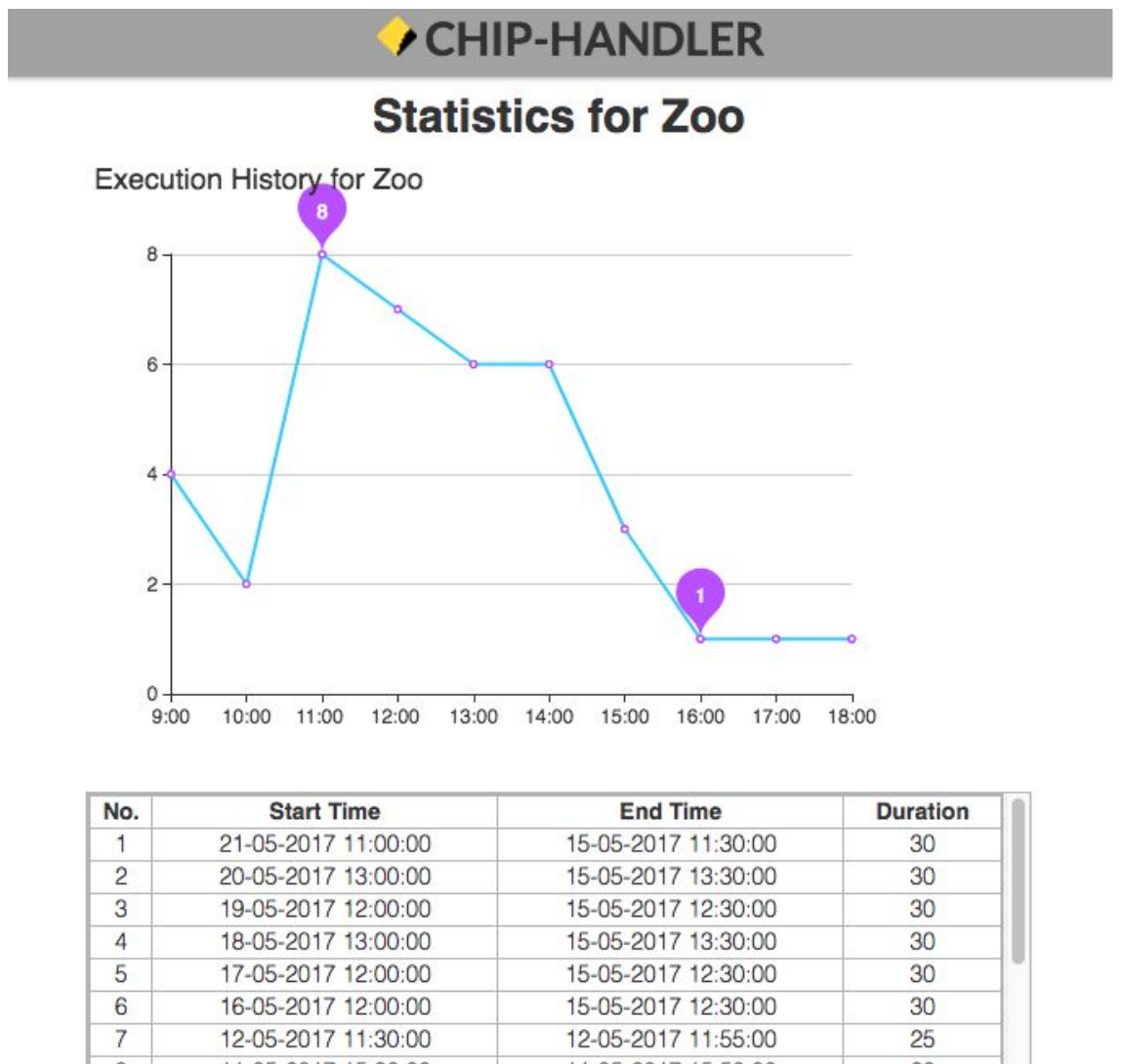}
\end{center}
\caption{Analysis of the data on the tour \emph{Zoo}}
\label{fig:stat2}
\end{figure}

Figure \ref{fig:stat1} presents the statistics on the  tours within the previous 6 months, as well as the distribution of he tours by their types, both in tabular and graphical format. 
The users can also obtain more detailed information on a particular tour, cf. Figure~\ref{fig:stat2}. 

\begin{figure}[ht!]
\begin{center} 
\includegraphics[width=7.5cm]{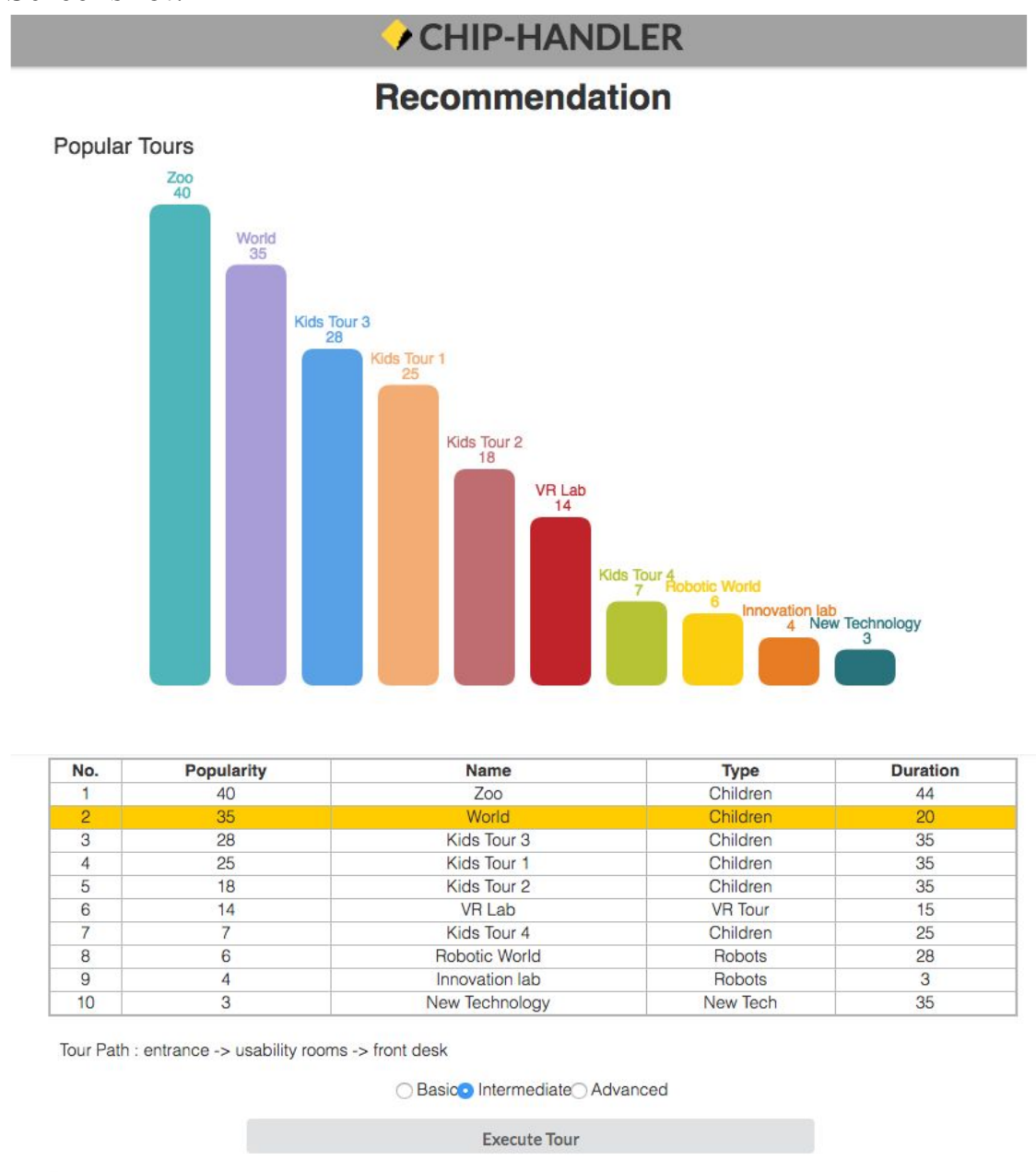}
\end{center}
\caption{Recommendations for the users based on the tour popularity}
\label{fig:rec1}
\end{figure}

Figure \ref{fig:rec1}  demonstrates how the developed system provides recommendations for the users based on the tour popularity. 
Customised recommendations, i.e., based on the customised parameters, are provided in a similar way.

To summarise, the developed web-based interface provides the following features:
\begin{itemize}
\item Navigate the robot,
\item Store the current locations of the robot,
\item Manage saved locations,
\item Create  a customised tour based on saved location,
\item Embed speeches for a customised tour,
\item Store information about a customised tour, such as the tour type, duration, etc.,
\item Provide recommendations for the users,  and
\item Visual analyse the data on previous tours.
\end{itemize} 

\newpage
\section{Discussion and Conclusions}
\label{sec:conclusions}

In this paper, we presented the core results of the project on  
software development for social robotics systems: 
We developed a web-based solution that supports the management of robot-guided tours (including the collection of the spatial information for the tours within noisy environments), provides recommendations for the users as well as allows for a visual analysis of the data on previous tours.
The results of our work were implement for the humanoid PAL REEM robot, but their core ideas can be applied for other types of humanoid robots.

We plan our future work on this project in three directions: 
\begin{itemize}
\item[(1)]
to embed  into the developed REEM framework the efficient testing methods, e.g. \cite{liu2015spatio,liu2015efficient,laali2016test}  as well as  the model-based hazard and impact analysis methods~\cite{dobi2013model},
\item[(2)]
to apply the developed framework to another type of ROS-based robot, Baxter, hosted in the RMIT University VXLab, and
\item[(3)]
to expand the developed guided tour features to involve game activities, as this would make the tours for children more entertaining and increase the children engagement.
\end{itemize}

\newpage
\section*{Acknowledgements} 
The project was sponsored by the Commonwealth Bank of Australia (CBA), Stockland Corporation Limited and the
Australian Technology Network of Universities (ATN). 
 We would like to thank William Judge (CBA) and Alec Webb (ATN) for numerous discussions and support.
 
\bibliographystyle{abbrv}

\end{document}